\pdfoutput=1

\documentclass[11pt]{article}

\usepackage{ACL2023}
\usepackage{booktabs}
\usepackage{hyperref}

\usepackage{times}
\usepackage{latexsym}
\usepackage{graphicx}

\usepackage{tablefootnote}
\usepackage{subcaption}
\usepackage{amsmath}
\usepackage{algorithm}
\usepackage{algorithmic}
\usepackage{listings}
\lstdefinelanguage{json}{
    basicstyle=\tiny\ttfamily,
    showstringspaces=false,
    breaklines=true,
    frame=lines,
    backgroundcolor=\color{white},
    literate=
      {:}{{{\color{blue}:}}}{1}
      {,}{{{\color{blue},}}}{1}
      {\{}{{{\color{blue}\{}}}{1}
      {\}}{{{\color{blue}\}}}}{1}
      {[}{{{\color{blue}[}}}{1}
      {]}{{{\color{blue}]}}}{1},
}

\usepackage[OT2,T1]{fontenc}
\newcommand\textcyr[1]{{\fontencoding{OT2}\fontfamily{wncyr}\selectfont #1}}

\usepackage[T1]{fontenc}

\usepackage[utf8]{inputenc}

\usepackage{microtype}

\usepackage{inconsolata}

\title{LLMs Are Zero-Shot Context-Aware Simultaneous Translators}

\author{Roman Koshkin$^{\dagger}$ \quad Katsuhito Sudoh$^{\ddagger \clubsuit}$ \quad Satoshi Nakamura$^{\ddagger \spadesuit}$ \quad \\
  $^\dagger$Okinawa Institute of Science and Tenchnology, Japan\\
  $^\ddagger$Nara Institute of Science and Technology, Japan \\
   $^\spadesuit$The Chinese University of Hong Kong, Shenzhen \\
   $^\clubsuit$Nara Women’s University, Japan \\
  \texttt{roman.koshkin@oist.jp} \\ \\
 \raisebox{-0.25\height}{\includegraphics[height=3.5ex]{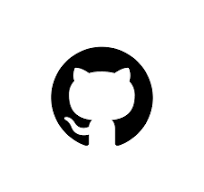}} \href{https://github.com/RomanKoshkin/toLLMatch}{\textcolor{red}{\texttt{https://github.com/RomanKoshkin/toLLMatch}}}
  }

\begin{document}

\maketitle

\begin{abstract}
The advent of transformers has fueled progress in machine translation. More recently large language models (LLMs) have come to the spotlight thanks to their generality and strong performance in a wide range of language tasks, including translation. Here we show that open-source LLMs perform on par with or better than some state-of-the-art baselines in simultaneous machine translation (SiMT) tasks, zero-shot. We also demonstrate that injection of minimal background information, which is easy with an LLM, brings further performance gains, especially on challenging technical subject-matter. This highlights LLMs' potential for building next generation of massively multilingual, context-aware and terminologically accurate SiMT systems that require no resource-intensive training or fine-tuning.
\end{abstract}

\section{Introduction}
In simultaneous translation, the translator -- either a machine or human -- is expected to start the translation \emph{before} the source sentence is finished, often making strong assumptions about the meaning of certain words, phrases, or the intent of the entire message. To produce a coherent -- although not necessarily accurate -- translation, human simultaneous translators routinely use a range of techniques, one of which is delaying the translation of an initially ambiguous word or phrase in the hope that its meaning will become resolved by later context \citep{Ilyukhin2001, chernov2004inference, setton2005pointing, AMOS2022104987}. Perhaps more importantly, human translators reduce this inherent uncertainty by relying on information from other sources, such as presentation slides and glossaries of standard terms. This, and the fact that some people insist on using the term "interpreter", rather than "translator"\footnote{Following the practice established in the machine translation community, in this paper we will be using the term "simultaneous translation".}, highlights a very different nature of this kind of translation. 

Despite significant progress in the field of offline machine translation, recently enabled by the wide adoption of the transformer architecture \citep{NIPS2017_3f5ee243}, the practical use of SiMT systems is still limited due to a range of unsolved problems. One of these problems is that existing SiMT systems -- in stark contrast to human simultaneous translators -- operate on a sentence level, completely disregarding the context established by previous sentences, or the broader (extralinguistic) context that is implied, but not contained in the text itself. Needless to say, such context-unaware translation is often logically incoherent and is prone to terminological inconsistencies, especially across long discourse. The very fact that human interpreters -- even the most experienced professionals -- routinely prepare for upcoming translation jobs by studying relevant subject-matter, reviewing or compiling topic-specific glossaries of terms, names, and job titles \citep{alvarez2022interpreter, gile1986travail, gile1985termes, Chernov1978}, suggests that SiMT systems should have access to \emph{additional} information needed to make terminologically appropriate and accurate translation.

\begin{figure*}[t]
  \centering
  \includegraphics[width=0.99\linewidth]{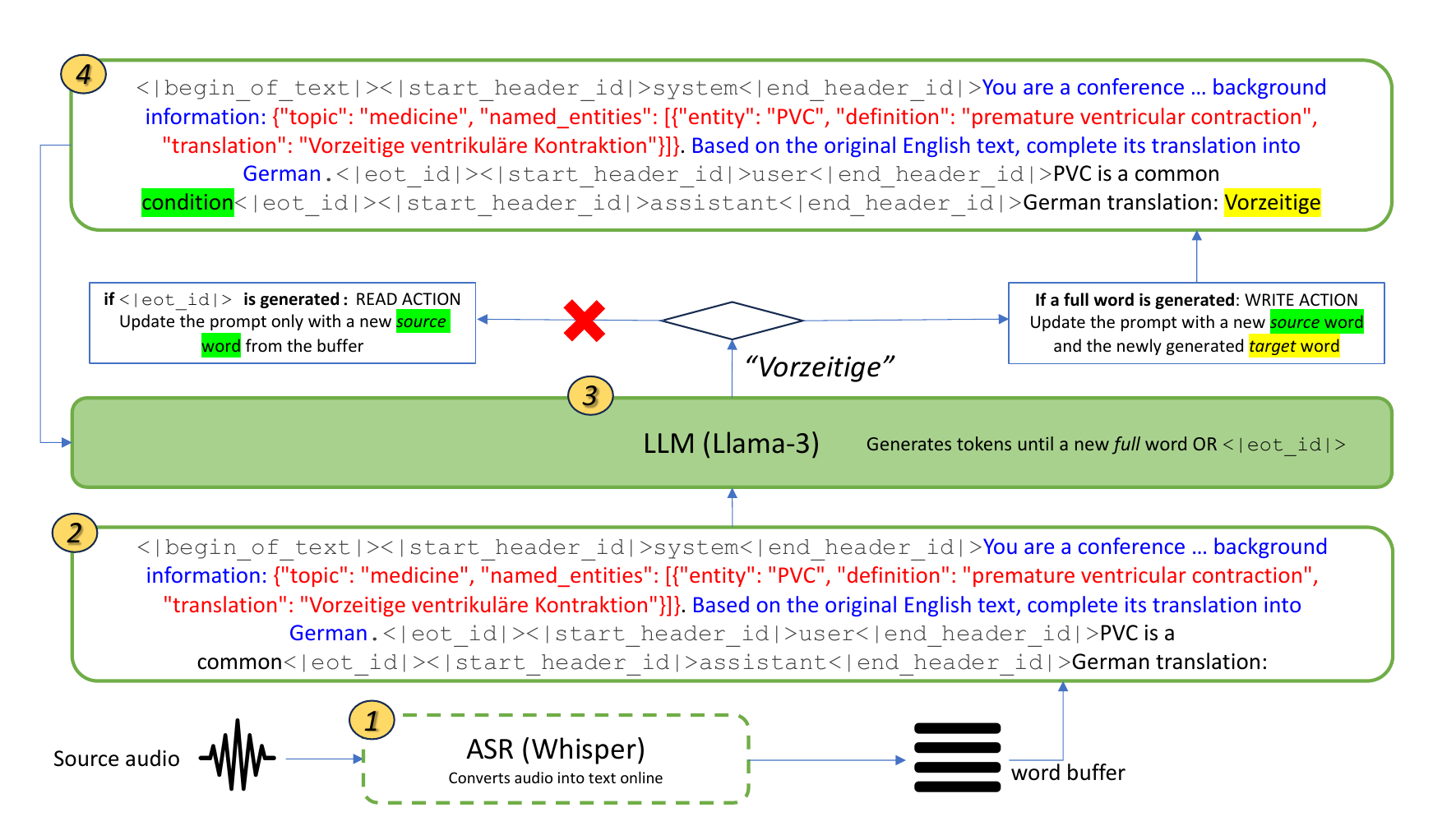}
   \caption{Model overview. Chunks of input audio are incrementally processed by \textsc{Whisper} (1), and the recognized words are stored in the buffer. The prompt (2) includes special strings (shown in grey), system message (blue) with background information (red) to constrain the space of possible translations, and the model's previous translation (if exists). Given the prompt, the LLM's generates tokens until either a new full word or \texttt{<|eot\_id|>} is generated (3). If a new full word is generated, a WRITE action is performed: a new source word from the word buffer and the newly generated word ("Vorzeitige" in this example) are  added to the prompt. If \texttt{<|eot\_id|>} is generated, a READ action is performed: the prompt is updated only with a new source word from the buffer.}
   \label{fig:hero}
\end{figure*}

Motivated by LLMs' strong reasoning \citep{yao2023tree, huang2024large, huang2023reasoning, zhou2024selfdiscover},  translation \citep{xu2024paradigm, zhu2023multilingual} and in-context learning \citep{liu-etal-2022-makes, wei2022chain, brown2020language} capabilities, we attempt to address one of the weaknesses of existing SiMT systems, namely that their translation takes no account of the wider context and generally cannot respect specific terminological constraints. Different from previous studies which have attempted fine-tuning LLMs for SiMT tasks \citep{wang2023simultaneous, agostinelli2023simul, koshkin2024transllama}, our focus here is on translation in zero-shot mode. In the method we propose, the LLM receives a prompt that contains \emph{both} the partial input, partial translation and minimal background information, and generates the next word of the translation. At the next step, the prompt is updated with the new source and the newly translated word (see Section \ref{sec:methods} for details). We show empirically that such an approach outperforms some of the strongest bilingual SiMT baselines and shows competitive results to a state-of-the-art multilingual SiMT system. Importantly, our approach makes it easy to insert background information (see Fig. \ref{fig:hero} and Section \ref{sec:experiments}), which helps the LLM to make \emph{contextually appropriate} word choices.

Our key contributions are as follows:

\begin{enumerate}
    \item We show that an off-the-shelf instruction-tuned LLM can successfully perform a SiMT task zero-shot, without a sophisticated segmentation policy, with quality and latency metrics that are competitive with (and in some cases exceeding) the state of the art.
    \item We show that instruction-tuned LLMs can be easily used for contextually-aware SiMT, and that injecting \emph{minimal} background information generally improves the quality of the translation by a large margin.
    \item We propose \emph{response priming}, which consists in fixing the initial part of the assistant's response, and improves the LLM's zero-shot performance on SiMT tasks.
      
\end{enumerate}

The rest of the paper is structured as follows. In Section \ref{sec:previous_work} we provide an overview of recent SiMT literature. In Section \ref{sec:methods} we describe our method and the datasets used for evaluating our method. In Section \ref{sec:experiments} we demonstrate the performance of our approach on the different datasets and language pairs. We conclude with a discussion of limitations and future directions and in Section \ref{limitations_and_future}.

\section{Related work}
\label{sec:previous_work}

Simultaneous machine translation (SiMT) systems strive to balance translation quality -- commonly evaluated using the BLEU metric \citep{papineni-etal-2002-bleu} -- with acceptable latency levels. This balance is managed through a "policy" that determines the timing of translation actions (i.e., a WRITE action) versus the reception of additional input (i.e., a READ action). The literature classifies these policies into two main types: fixed and adaptive \citep{zhang-etal-2020-learning-adaptive}. Fixed policies, such as \emph{wait-k} \citep{ma-etal-2019-stacl}, apply predefined rules for executing READ and WRITE actions, regardless of the textual context. Initially, SiMT models employed \emph{chunk-based} strategies \citep{bangalore2012real, yarmohammadi2013incremental,fugen2007simultaneous, sridhar2013segmentation}, where the text is divided into sub-sentence segments for translation without considering the context from preceding chunks, leading to reduced translation accuracy. In response to these drawbacks, \citet{dalvi-etal-2018-incremental} introduced an \emph{incremental decoding} method. This technique enhances chunk translations by integrating preceding contexts via the hidden states of an RNN. Paired with straightforward segmentation tactics, their method surpassed the performance of prior state-of-the-art systems. Meanwhile, adaptive policies, such as "wait-if" rules \citep{cho2016neural}, allow for more flexible WRITE/READ actions by considering parts of the source and/or target text. Adaptive policies can be developed using separately trained agents, often employing reinforcement learning techniques \citep{alinejad-etal-2018-prediction, Satija2016SimultaneousMT, grissom-ii-etal-2014-dont, gu-etal-2017-learning}. These policies may initiate READ/WRITE actions based on model attention mechanisms \citep{Ma2020Monotonic, arivazhagan-etal-2019-monotonic, 10.5555/3305890.3305974, DBLP:conf/iclr/ChiuR18} or the stability of output predictions across $n$ steps, a concept referred to as "local agreement" \citep{polak-etal-2022-cuni, ko-etal-2023-tagged, liu20s_interspeech}. Recent research has also investigated policy training using binary search strategies \citep{guo-etal-2023-learning} to optimize the translation quality improvement per token processed, and has conceptualized the translation actions as a hidden Markov transformer \citep{zhang2023hidden}, where hidden events indicate optimal translation output times.

A promising area of research, related to this study, focuses on adapting encoder-decoder transformers like mBART \citep{liu-etal-2020-multilingual-denoising}, which were initially developed for sentence-level translation, to the SiMT task. For instance, \citet{kano-etal-2022-simultaneous, fukuda-etal-2023-naist} have applied fine-tuning techniques using prefix-alignment data, while \citet{zhang-etal-2020-learning-adaptive} have employed fine-tuning on "meaningful units", both demonstrating strong performance across various language pairs.

More recently, large language models (LLMs) have demonstrated remarkable capabilities across a wide range of tasks, including offline machine translation \citep{xu2024paradigm, zhu2023multilingual}. Importantly, LLMs' ability to learn in-context enables a range of new capabilities, such as terminology-constrained translation \citep{moslem2023adaptive} and self-correction of translation errors \citep{feng2024improving}. These and other developments raised the question whether LLMs can be leveraged for SiMT. Recent works have explored various ways to fine-tune LLMs for SiMT and showed that coupled with a segmentation policy, such as wait-k \citep{wang2023simultaneous} or more sophisticated "local agreement" \citep{agostinelli2023simul}, it can deliver competitive performance on some language pairs. \citet{koshkin2024transllama} proposed a policy-free approach, in which an LLM is fine-tuned on pairs of "causally aligned" source-target sentence pairs to act as both the translator and segmentation policy at the same time.

Distinct from previous literature, we show that an off-the-shelf instruction-tuned LLM can perform SiMT zero-shot, eliminating the need for resource-intensive model training and the complexities of making special datasets and fine-tuning. Importantly, our approach enables \emph{context-aware SiMT} which, as we empirically demonstrate, substantially improves translation quality.

\section{Method}
\label{sec:methods}

\subsection{Online ASR} 

Similarly to \citet{koshkin2024transllama}, we follow a cascaded approach, where an automatic speech recognition (ASR) model (\textsc{Whisper} \citep{radford2023robust}) incrementally converts input audio chunks into text which is fed into the LLM for translation. We found that for English input \texttt{whisper-small.en}\footnote{\texttt{https://huggingface.co/openai/whisper-small.en}} achieved approximately the same word error rate (WER) of about about 5\% as \texttt{whisper-large-v3}, so we chose the smaller version for faster inference. Although trained on full sentences, \textsc{Whisper} can still perform online ASR with the following simple technique. For each READ action, a new segment of audio, lasting 200 ms, is added to any previously read audio chunks and then processed by \textsc{Whisper}. This window length was chosen empirically as a trade-off between, on the one hand, the desire to minimize translation latency and word error rate (WER): larger windows typically are likely to result in lower WER, but tend to increase latency metrics. In our online ASR, we discard the last predicted word unless the entire source audio has been read in.

Similarly to \citet{koshkin2024transllama}, the output of the ASR cascade is fed into the LLM (\texttt{Llama-3-70B-Instruct\footnote{At the time of writing this paper, Meta had released the 8B and 70B versions of the model, but not the corresponding paper or technical report.}}). However, in an important distinction from \citet{koshkin2024transllama}, we insert the partial target not into the "user", but the "assistant" part of the prompt (Fig. \ref{fig:hero}). This simple modification, which we call \emph{response priming}, effectively limits the space of possible sequences that the model can produce and prevents it from generating apologies, explanatory notes or other undesirable additions to the translation.

\subsection{Evaluation Data}
\label{data}

For the English-German language pair we used \textsc{Fleurs} \citep{conneau2022fleurs} and \textsc{TED-TST-2023} \citep{koshkin2024transllama}. However, it is possible that those test sets (or the data that they were built from) were leaked into the LLM's pre-training set. For this reason we created another dataset -- which we call \textsc{TED-TST-2024} -- similar in size and content type to \textsc{TED-TST-2023}, but only including talks posted after the LLM was released.

Additionally, to showcase the ability of LLMs to leverage background information for improved SiMT, we context-augment \textsc{TED-TST-2023} and \textsc{TED-TST-2024} with relevant background information (Listing 1).

\begin{lstlisting}[language=json, caption={Example of background information used to augment \textsc{TED-TST-2023} and \textsc{TED-TST-2024}.}]
{
    "topic": "Climate Crisis and Fossil Fuel Industry's Influence", 
    "named_entities": 
        [ 
            {"entity": "troposphere", 
            "description": "the lowest part of the atmosphere"}, 
            {"entity": "Inflation Reduction Act", 
            "description": "U.S. legislation aimed at addressing climate change"},
            {"entity": "COP process", 
            "description": "Conference of the Parties, climate change conferences"}, 
            {"entity": "COP28", 
            "description": "upcoming climate conference hosted by UAE"},
        ]
}
\end{lstlisting}
\label{listing1}

We generated this background information with \texttt{gpt-4-turbo-2024-04-09} by prompting it with the entire TED talk for which a given sentence was taken (the full prompt is in Appendix \ref{extraction_prompt}). The idea here is to make the translation more realistic by providing the translator (the LLM in our case) with essential information about the subject-matter at hand. 

Finally, we test our model in a more challenging scenario imitating translation of highly technical subject-matter. Prior to translating complex, technical subject matter, human interpreters compile topic-specific glossaries, which typically list terms from the source language along with their definitions and standard translations into the target language \citep{alvarez2022interpreter, gile1986travail, gile1985termes, Chernov1978}. This preparatory work is crucial for effectively conveying technical content, as it equips interpreters with the precise terminology and contextual knowledge needed to handle subject-specific nuances. Motivated by this, we constructed \textsc{AmbiEval}, which is a context-augmented dataset of ambiguous terms, which we describe next. First we collect a list of English words (some of which are acronyms) that can have very different meanings in different contexts. For example, depending on the context, the word "MOS" can mean "metal oxide semiconductor" and also "military occupational specialty". Sometimes, the meaning of the word is disambiguated later in the sentence. Consider the following two examples:

\emph{One must watch out for kicks, which are dangerous influxes of formation fluids into the wellbore.}

\emph{One must watch out for kicks, while maintaining a strong defense and executing effective strikes.}

In these sentences, the meaning of the word "kicks" is disambiguated by later context, specifically by the words "influxes" and "strikes". Unless background information is somehow fed into the model together with the source, it is difficult for the SiMT model to immediately translate the word "kicks" accurately. We also create examples with words whose meaning cannot be disambiguated based on the information contained within the sentence, for example:

\emph{The CPA recommends holding pharmaceutical companies to stricter standards of accountability.}

In this sentence, "CPA" is never disambiguated and can mean almost anything (e.g. "Consumer Protection Act", "Canadian Psychiatric Association", "Cerebral Palsy Alliance"). The source audio of \textsc{AmbiEval} is generated by Amazon's Polly text-to-speech service.

\subsection{Inference} 

For inference, we follow a similar approach to \textsc{TransLLama} \citep{koshkin2024transllama}, but also inject background information. Specifically, at time $t$, the target token $y_t$ is conditional on all the source tokens $x_{\leq t}$ revealed up to time $t$,  previously generated target tokens $x_{<t}$ and background information $b$, which is constant for sentences coming from the same text (speech).

\begin{equation}
    p(y_{t}|y_{<t}, x_{\leq t}, b)
\label{eq:loss}
\end{equation}

Given a prompt (Fig. \ref{fig:prompt}) consisting of a system message, partial input and previously translated partial target, the LLM greedily generates one or more new tokens. Once a new full word is generated, a WRITE action is performed. A READ action is performed when an \texttt{<|eot\_id|>} token is generated. A WRITE action involves adding the next source word and the newly translated target word to the prompt. In a READ action, the prompt is only updated by inserting the next source word into the prompt. WRITE actions are only permitted after the length of the input audio reaches a certain minimum length. This constraint controls latency-quality trade-off and indirectly the WER: higher values of this minimum length generally improve the quality by increasing the average number of words the LLM gets at the beginning of translation and decreasing the WER of the ASR\footnote{if the initial audio segment is too short, \textsc{Whisper} is more likely to hallucinate words that were never said.}. Except for the temperature (set to 0 for greedy generation), all the generation parameters were left at their default values.

\begin{figure*}[ht]
  \centering
  \begin{verbatim}
  <|begin_of_text|><|start_header_id|>system<|end_header_id|>
  SYSTEM_MESSAGE
  BACKGROUND_INFORMATION_JSON
  USER_INSTRUCTION
  <|eot_id|><|start_header_id|>user<|end_header_id|>
  Context: PARTIAL_SOURCE
  <|eot_id|><|start_header_id|>assistant<|end_header_id|>
  German translation: PARTIAL_TARGET
  \end{verbatim}
  
  \caption{Prompt structure. \texttt{<|begin\_of\_text|>, <|start\_header\_id|>ROLE\_NAME<|end\_header\_id|>}, and \texttt{<|eot\_id|>} are special strings used in Llama-3 to flank the system, user and assistant parts of the the prompt.}
  \label{fig:prompt}
\end{figure*}

After all the source words have been revealed, the input is no longer partial and no new words are added to it, but the generation process continues until \texttt{<EOS>}. We illustrate the inference process in Fig. \ref{fig:hero} and Algorithm \ref{alg:cap0}.

\begin{algorithm}
\caption{Inference process}
\label{alg:cap0}

\begin{lstlisting}
partial_output = []

# do ASR after MIN_T s of audio is read
asr = ASR(min_t=MIN_T)
llm = LLM()

while True:
    # get the next audio chunk, recognize
    (partial_input, 
    audio_finished) = asr.next()
    
    prompt = " ".join([
        SYSTEM_MSG, 
        background_info,
        partial_input, 
        partial_output])

    # generate until full word 
    # or `<|eot_id|>`
    next_word = llm.generate(prompt)

    if next_word == "<|eot_id|>":
        if audio_finished:
            break    # finish sentence
        else:
            continue # READ
    else:
        # WRITE
        partial_out.append(
            next_word) 
\end{lstlisting}

\end{algorithm}

For fast inference, we use the \texttt{vllm}\footnote{\texttt{https://github.com/vllm-project/vllm}} library which implements a range of latest LLM performance optimizations, most importantly tensor parallelism. 
Unless otherwise noted, all the results reported in this paper were obtained on a Linux machine with 4 A100 80GB GPUs. The ASR cascade was run using \texttt{whisper-jax}\footnote{\texttt{https://github.com/sanchit-gandhi/whisper-jax}} an implementation of \textsc{Whisper} built for maximum inference speed.

\subsection{Prompt structure} 

We follow a similar prompt structure as in \citet{koshkin2024transllama} (Fig. \ref{fig:prompt}), except that we do not instruct the LLM to generate special \texttt{<WAIT>} tokens, but inject background information as part of the system message. For the \texttt{SYSTEM\_MESSAGE} we used the following text: \emph{"You are a conference interpreter. As you translate, you can use the following background information: \texttt{BACKGROUND\_INFORMATION\_JSON}. Taking into account the original \texttt{SRC\_LANG} text, complete its translation into \texttt{TGT\_LANG}. Do not add any notes or comments to the translation."} This system message performed well empirically, and we speculate that further improvements are possible with different system messages. We leave this question to future work.

\section{Results}
\label{sec:experiments}

\subsection{Benchmarks}

In this section we compare the performance of our method to \textsc{SeamlessStreaming} \citep{barrault2023seamless}, which is a state-of-the-art massively multilingual SiMT system on five language pairs (\texttt{en-\{de,es,fr,it,ru\}}) and additionally to three recent bilingual SiMT systems, namely: \textsc{NAIST} \citep{fukuda-etal-2023-naist}, \textsc{FBK} \citep{papi-etal-2023-attention} and \textsc{TransLLaMa}\footnote{We used the version of \textsc{TransLLaMA} derived from Llama-2-70B.} \citep{koshkin2024transllama} on the \texttt{en-de} pair.


\begin{figure*}[h]
  \centering
  \includegraphics[width=\linewidth]
  {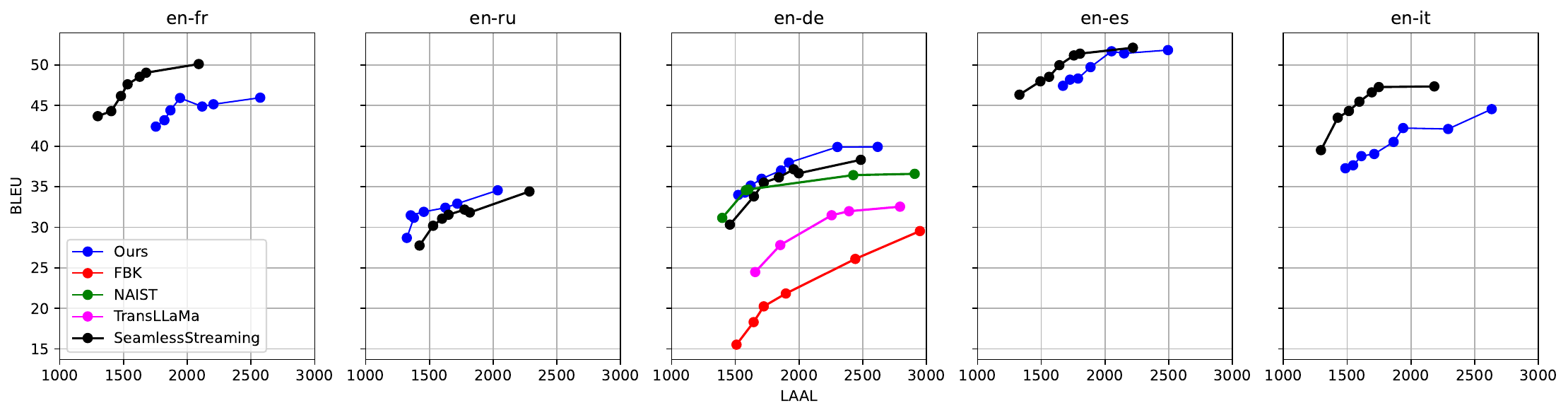}
  \caption{Dependence of translation quality (measured by BLEU) on latency (measured by LAAL) for \texttt{en-\{fr, ru, de, es, it\}} on \textsc{TED-TST-2024}. The latency was controlled by varying the minimum length of the audio before allowing WRITE actions (\textsc{Ours} and \textsc{TransLLaMa}), attention threshold (\textsc{SeamlessStreming} and \textsc{FBK}) and source segment size (\textsc{NAIST}).}
  \label{fig:Fig1}
\end{figure*}

We start by examining the quality-latency tradeoff on \textsc{TED-TST-2024} (Fig. \ref{tab:TED_baselines}). Our method performed strongly relative to the recent baselines (although not on all language pairs). In all of the results presented in this section, we controlled the translation latency by varying the minimum length of the audio before allowing WRITE actions (\textsc{Ours} and \textsc{TransLLaMa}), attention threshold (\textsc{SeamlessStreming} and \textsc{FBK}) and source segment size (\textsc{NAIST}). When benchmarking our model against the baselines on the \textsc{Fleurs}, \textsc{TED-TST-2023}, and \textsc{AmbiEval} datasets, we approximately matched the length-aware average lagging (LAAL) \citep{papi-etal-2022-generation} to 2000 ms.

\begin{table}[h]
\centering
\begin{tabular}{lrrll}
\toprule
Method & BLEU & AL & LAAL \\
\midrule
Ours & 22.13 & 1360.59 & 2089.16 \\
\textsc{NAIST} & 21.39 & 1060.94 & 1967.36  \\
\textsc{FBK} & 17.65 & 1645.42 & 1922.79  \\
\textsc{Seamless} & 19.75 & 1442.71 & 1781.06 \\
\textsc{TransLLaMA} & 19.36 & 1732.08 & 2017.91 \\
\bottomrule
\end{tabular}
\caption{Quality and latency results for our approach compared with state-of-the-art baselines on the \texttt{en-de} language pair on \textsc{TED-TST-2023}.}
\label{tab:TED_baselines}
\end{table}

Additional performance tests on \textsc{TED-TST-2023} (Table \ref{tab:TED_baselines}) and \textsc{Fleurs} (Table \ref{tab:fleurs}) further demonstrate the performance of our approach. Since \textsc{TED-TST-2023} and \textsc{TED-TST-2024} are built from content intended for lay audiences, and therefore is relatively easy to translate, we also evaluate our method on another dataset (\textsc{AmbiEval}) which models a more challenging scenario where the meaning of some technical terms cannot be resolved immediately or without additional contextual information (see Section \ref{data}). As expected, our method outperforms the baselines by a large margin (Table \ref{tab:AmbiEval}, but also see Section \ref{sec:ablations}).

\begin{table}[h]
\centering
\begin{tabular}{lrrll}
\toprule
Method & BLEU & AL & LAAL \\
\midrule
Ours & 32.30 & 1720.00 & 2022.05\\
\textsc{NAIST} & 36.44 & 1615.80 & 2120.09\\
\textsc{Seamless} & 31.75 & 1695.24 & 1877.11 \\
\textsc{FBK} & 15.56 & 1744.59 & 2028.93 \\
\textsc{TransLLaMA} & 25.71 & 1820.33 & 2095.07 \\

\bottomrule
\end{tabular}
\caption{Quality and latency results for our approach compared with state-of-the-art baselines on the \texttt{en-de} language pair on \textsc{Fleurs}.}
\label{tab:fleurs}
\end{table}

\begin{table}[h]
\centering
\begin{tabular}{lrrll}
\toprule
Method & BLEU & AL & LAAL \\
\midrule
Ours & 42.60 & 1961.57 & 2008.48  \\
\textsc{FBK} & 24.96 & 1906.59 & 2151.32  \\
\textsc{NAIST} & 39.80 & 1662.06 & 1796.68  \\
\textsc{Seamless} & 29.76 & 1937.35 & 1978.72  \\
\textsc{TransLLaMA} & 32.43 & 1838.81 & 1903.21 \\
\bottomrule
\end{tabular}
\caption{Quality and latency results for our approach compared with state-of-the-art baselines on the \texttt{en-de} language pair on \textsc{AmbiEval}.}
\label{tab:AmbiEval}
\end{table}

\subsection{Inference speed}

One might wonder if using an LLM for real-time SiMT is feasible in practice. While our system has much more parameters than the state-of-the-art SiMT baselines (except for \textsc{TransLLaMa}), it can still achieve real-time translation if run on a modern inference engine that leverages a range of optimizations such as tensor parallelism (Table \ref{tab:param_RTF}).

\begin{table}[h]
\centering
\begin{tabular}{lrrr}  
\toprule
Method & bn params & RTF \\
\midrule
Ours & 70.79 & 0.86  \\
\textsc{NAIST} & 1.04 & 1.34 \\
\textsc{FBK} &  0.176 & 0.42 \\  
\textsc{Seamless} & 1.96 & 0.36  \\
\textsc{TransLLaMa} & 70.52\tablefootnote{Assuming \texttt{whisper-large-v2} is used for ASR.} & 15.3  \\
\bottomrule
\end{tabular}
\caption{Parameter counts and real-time factor  (RTF) of the chosen baselines and our model. See Appendix \ref{rtf} for information about model hyperparameters and how RTF was calculated.}
\label{tab:param_RTF}
\end{table}

As long as the entire system -- including the ASR cascade and LLM -- can function with an RTF of 1 or less, it can in principle be used for live simultaneous translation.

\subsection{Recovery from ASR errors}
Beyond the ability to ingest additional (background) information, another advantage of LLM-based translation is the ability to recover from ASR errors \citep{NEURIPS2023_64922674, hu2024listen, Yang_2023, ma2023generative}. Although on the TED datasets \textsc{Whisper} produces a very low WER ($<5\%$), these errors might still negatively impact the translation quality. Inspection of the translated texts reveals that compared to a state-of-the art offline translation model (\textsc{NLLB-200} \citep{nllbteam2022language}) Llama-3 is very good at correcting ASR errors, for example:

\textbf{ASR output:} \emph{I think terrorists like Hamas and \underline{his bala} are evil, and there is a bright line between groups that aim to kill \underline{innocence} and those that try to avoid doing so at all costs.}

\textbf{LLM translation:} \emph{Ich denke , Terroristen wie Hamas und Hezbollah sind böse, und es gibt eine klare Grenze zwischen Gruppen, die unschuldige Menschen töten wollen, und jenen, die alles tun, um dies zu vermeiden.}

\textbf{NLLB translation:} \emph{Ich denke, Terroristen wie die Hamas und seine Bala sind böse, und es gibt eine klare Linie zwischen Gruppen, die Unschuld töten wollen, und denen, die versuchen, dies um jeden Preis zu vermeiden.}

In the example above, two ASR errors (underlined in the ASR output) were corrected by the LLM, but not by \textsc{NLLB-200}. For more examples, see Appendix \ref{examples_of_asr_correction}.

\subsection{Ablations}
\label{sec:ablations}

\textbf{Response priming}. Table \ref{tab:priming_ablation} shows that removing response priming from the prompt results in a small but consistent decrease of translation quality. This makes sense because response priming constrains the space of possible sequences that the LLM can generate in response to the prompt. Inspection of the translations revealed that without response priming the translations often begin with unwanted notes, comments and explanations resulting in decreased quality.

\begin{table}[ht]
\centering
\small
\begin{tabular}{lccccc}
\toprule
priming & en-de & en-es & en-fr & en-it & en-ru \\
\midrule
yes & 41.43 & 54.87 & 47.21 & 40.24 & 36.38 \\
no & 39.52 & 54.53 & 46.06 & 38.71 & 36.11 \\
\bottomrule
\end{tabular}
\caption{Disabling response priming consistently decreases translation quality across all the five language pairs. The numbers are mean BLEU scores over five runs with different latencies on \textsc{TED-TST-2024}.}
\label{tab:priming_ablation}
\end{table}

\textbf{Background information}. The removal of minimal background information notably decreases the translation quality (Table \ref{tab:background_ablation}), highlighting that the LLM can leverage even minimal information for improved quality. Notably, the smaler version of \textsc{Llama-3} does not seem to benefit from added background information (Table \ref{tab:8B_ablation}), which is likely due to the fact that smaller LLMs generally have weaker instruction-following and in-context learning abilities.

\begin{table}[ht]
\centering
\small
\begin{tabular}{lrrrrr}
\toprule
background & en-de & en-es & en-fr & en-it & en-ru \\
\midrule
no & 31.14 & 46.04 & 41.76 & 36.38 & 29.11 \\
yes & 36.76 & 49.81 & 44.57 & 40.26 & 31.87 \\
\bottomrule
\end{tabular}
\caption{Removing background information from the prompt significantly and consistently decreases quality across the all the five language pairs. The numbers are mean BLEU scores over five runs with different latencies on \textsc{TED-TST-2024}.}
\label{tab:background_ablation}
\end{table}

\textbf{Smaller LLMs}. Is is possible to achieve comparable performance (in terms of quality) with a smaller LLM? Our tests show that, unfortunately, \texttt{Meta-Llama-3-8B-Instruct} significantly underperforms its larger version, \texttt{Meta-Llama-3-70B-Instruct} and seems to be unable to benefit from background information (Table \ref{tab:8B_ablation}). Inspection of the translations suggests that the the smaller LLM is much worse at exactly following the instruction to only output the translation and nothing else.

\begin{table}[ht]
\centering
\small
\begin{tabular}{llrrr}
\toprule
background & pair & BLEU & AL & LAAL \\
\midrule
 & en-de & 30.52 & 2311.31 & 2466.86 \\
 & en-fr & 41.91 & 2609.47 & 2678.53 \\
yes & en-es & 41.76 & 2520.15 & 2626.96 \\
 & en-ru & 26.14 & 2018.06 & 2254.75 \\
 & en-it & 31.76 & 2356.28 & 2567.44 \\
 \hline
 & en-de & 30.42 & 2313.28 & 2404.13 \\
 & en-fr & 41.96 & 2621.87 & 2691.50 \\
no & en-es & 42.79 & 2519.84 & 2605.44 \\
 & en-ru & 26.40 & 2025.78 & 2226.59 \\
 & en-it & 36.23 & 2357.07 & 2454.95 \\
\bottomrule
\end{tabular}
\caption{A smaller LLM performs significantly worse than the default 70B version. Results are shown for the \textsc{TED-TST-2024} dataset.}
\label{tab:8B_ablation}
\end{table}

\section{Limitations and Future Directions}
\label{limitations_and_future}

Prior work has demonstrated that fine-tuning on a small dataset is sufficient to enable an LLM to perform the challenging task of simultaneous translation. However, these existing approaches are potentially limited to one language pair, involve constructing a specialized dataset and a non-trivial search for optimal fine-tuning hyperparameters. Here we demonstrate the an off-the-shelf instruction-tuned LLM performs strongly zero-shot on several different datasets and, crucially, can leverage additional information for improved quality and/or adherence to a predefined list of technical terms, which is important in translating technical material. 

In the future, as stronger and more lightweight models become available, the LLM can analyze its own translations and/or summarize source sentences or paragraphs. These summaries could be added to a vector store or a graph database and retrieved in real time to augment the translation of future sentences.

The big performance gap between the 8B and 70B version of \textsc{Llama-3} suggests that even better translation quality could be achieved with larger closed-source models (such as \textsc{GPT-4} or \textsc{Claude}) if their APIs allowed response priming.

One practical limitation of our approach is that currently, to the best of our knowledge, it cannot be used with strong closed-source models that are available through API. Perhaps as a countermeasure against model jailbreaking, the APIs through which these instruction-tuned models (e.g. GPT-4, Claude and Gemini) can be accessed enforce a rigid prompt structure that is incompatible with \emph{response priming} -- specifying a user-specified prefix for the (assistant) model's response -- which is at the core of our approach.

Another significant bottleneck in our LLM-based simultaneous translation system is that it relies on a separate ASR system that was not designed for online operation. Although in general this cascaded setup works well, hallucinations sometimes occur, especially in low-latency regimes when in response to initial silence \textsc{Whisper} outputs words that were never said in the audio. We believe this limitation can be addressed by implementing an end-to-end SiMT system, in which the output embeddings of an ASR system or speech encoder  would be directly projected into the LLM's input embedding space, bypassing a text representation and improving the system's latency overall. In fact, there is already some work in this direction, e.g. by \citet{fathullah2023prompting} and \citet{huang2023speech}.
 
It is interesting to explore other ways to improve the performance and efficiency of our method, such as local agreement \citep{polak-etal-2022-cuni}, efficient weight quantization (e.g. \texttt{awq} \citep{lin2023awq}), and more sophisticated prompting strategies.

\bibliography{anthology,custom}
\bibliographystyle{acl_natbib}

\clearpage
\appendix

\section*{Appendix}

\section{Prompts}
\label{extraction_prompt}

\vspace{10pt}
Prompt used to extract background information for \textsc{TED-TST-2023} and \textsc{TED-TST-2024}:
\vspace{10pt}

\begin{lstlisting}[language=json]

Please extract the topic and named entities (which are either proper names, technical terms or acronyms) from the following text, and return them as a JSON object with the following fields: topic, named_entities({entity, description}). For example:

{
  "topic": "Climate Crisis and Fossil Fuel Industry's Influence",
  "named_entities": [
    {
      "entity": "troposphere",
      "description": "the lowest part of the atmosphere"
    },
    {
      "entity": "gravity-measuring satellite",
      "description": "satellite used to observe water surpluses and deficits"
    },
    {
      "entity": "Inflation Reduction Act",
      "description": "U.S. legislation aimed at addressing climate change"
    },
    {
      "entity": "fossil fuel industry",
      "description": "industry opposing climate legislation"
    },
    {
      "entity": "UN Secretary General",
      "description": "stated fossil fuel industry is the 'polluted heart' of climate crisis"
    },
    {
      "entity": "COP process",
      "description": "Conference of the Parties, climate change conferences"
    },
    {
      "entity": "COP28",
      "description": "upcoming climate conference hosted by UAE"
    },
    {
      "entity": "Sultan Al Jaber",
      "description": "CEO of Abu Dhabi National Oil Company and president of COP28"
    },
    {
      "entity": "Paris Agreement",
      "description": "international treaty on climate change"
    },
    {
      "entity": "Chevron",
      "description": "company with 93% worthless and junk offsets"
    },
    {
      "entity": "World Bank",
      "description": "institution needing reform to aid developing countries"
    },
    {
      "entity": "gigafactory",
      "description": "large factory for producing batteries and renewable energy components"
    },
    {
      "entity": "Drew Shindell",
      "description": "researcher on CO2 reduction timelines"
    }
  ]
}

TEXT
\end{lstlisting}

\section{Examples of ASR error correction}
\label{examples_of_asr_correction}

\vspace{10pt}
\emph{Example 1}
\vspace{10pt}

\textbf{ASR output:} Our mistaken belief that the best leaders are those who generate the most noise, action, and sensational activity in the most \underline{traumatic} circumstances.

\textbf{LLM translation:} Unsere irrige Annahme, dass die besten Führer jene sind, die den meisten Lärm, die meisten Aktionen und die spektakulärsten Aktivitäten in den dramatischsten Umständen erzeugen.

\textbf{Offline MT (NLLB):} Unsere falsche Überzeugung, dass die besten Führungskräfte sind, die die meisten Lärm, Aktion und sensationelle Aktivität in den traumatischsten Umständen erzeugen.

\vspace{10pt}
\emph{Example 2}
\vspace{10pt}

\textbf{ASR output:} In contrast, the expeditions of the \underline{role} Amundsen make for boring reading.

\textbf{LLM translation:} Im Gegensatz dazu stehen die Expeditionen des Roald Amundsen für langweilige Lektüre.

\textbf{Offline MT (NLLB):} Im Gegensatz dazu sind die Expeditionen der Rolle Amundsen für langweiliges Lesen.

\vspace{10pt}
\emph{Example 3}
\vspace{10pt}

\textbf{ASR output:} If Shackleton is the swimmer who rushes recklessly into the water without understanding the currents or \underline{its} own capabilities, amundsen is the swimmer who has spent a lifetime, humbly studying the river before entering the water in just the right spot, at just the right time and so makes it look easy.

\textbf{LLM translation:} Wenn Shackleton der Schwimmer ist, der sich ohne vorheriges Verständnis der Strömungen oder seiner eigenen Fähigkeiten tollkühn ins Wasser stürzt, dann ist Amundsen der Schwimmer, der ein Leben lang bescheiden den Fluss studiert hat, bevor er genau am richtigen Punkt und zur richtigen Zeit ins Wasser geht und es dadurch leicht aussehen  lässt.

\section{Examples of translation}
\label{example_translations}

\vspace{20pt}
English source: I would like to invite you on a little thought experiment.

German translation: Ich möchte Sie einladen zu einer kleinen Gedankenübung.

Russian translation: \textcyr{Я хотел бы пригласить вас на маленький мысленный эксперимент.}

Italian translation: Vorrei invitarti a una piccola riflessione immaginativa.

French translation: Je voudrais vous inviter à une petite expérience de pensée.

Spanish translation: Me gustaría invitarle a una pequeña experiencia de pensamiento.

\vspace{20pt}
English source: Let's pretend that we're going on a polar expedition together, all of you and me, and we need to hire a captain.

German translation: Lassen Sie uns so tun, als würden wir gemeinsam eine Polarexpedition unternehmen, alle zusammen, Sie und ich, und wir müssen einen Kapitän einstellen.

Russian translation: \textcyr{Давайте представим, что мы отправляемся в полярную экспедицию вместе, все вы и я, и нам нужно нанять капитана.}

Italian translation: Immaginiamo di essere in una spedizione polare insieme, tutti voi e io, e dobbiamo assumere un capitano.

French translation: Imaginons que nous partions ensemble pour une expédition polaire, vous et moi, et que nous devions embaucher un capitaine.

Spanish translation: Imaginemos que estamos embarcándonos en una expedición polar juntos, todos ustedes y yo, y necesitamos contratar a un capitán.

\vspace{20pt}
English source: One comes from a man who has already successfully achieved all four of the major polar goals, the North Pole and the South Pole, and the Northeast and the Northwest Passage.

German translation: Einer stammt von einem Mann, der bereits alles vier erfolgreich erreicht hat, was die großen polaren Ziele betrifft, den Nordpol und den Südpol sowie die Nordost- und die Nordwestpassage.

Russian translation: \textcyr{Один от мужчины, который уже успешно достиг все четырех полярных целей: Северного полюса и Южного полюса, а также Северо-восточного и Северо-Западного проходов.}

Italian translation: Uno proviene da un uomo che ha già conseguito con successo tutti e quattro gli obiettivi polari maggiori, il Polo Nord e il Polo Sud e il Passaggio del Nordest e il Passaggio del Nordovest.

French translation: L'un vient de l'homme qui a déjà réussi à tout accomplir dans les quatre grands objectifs polaires, le Pôle Nord et le Pôle Sud, ainsi que le Passage du Nord-Est et le Passage du Nord-Ouest.

Spanish translation: Uno proviene de un hombre que ya ha logrado con éxito los cuatro objetivos polares principales, el Polo Norte y el Polo Sur , así como el Paso del Noreste y el Paso del Noroeste.

\vspace{10pt}
English source: In fact, three of these, he was the first person to accomplish.

German translation: Tatsächlich drei von ihnen war er der Erste, der dies erreicht hat.

Russian translation: \textcyr{Фактически, три из них он был первым человеком, который это совершил.}

Italian translation: In realtà, tre di questi, fu la prima persona a realizzare.

French translation: En réalité, trois d'entre eux, il fut le premier à accomplir.

Spanish translation: De hecho, tres de ellos, fue la primera persona en lograr.

\vspace{20pt}
English source: Candidate B is a man who set off for the Antarctic four times, three times as the man in charge, and every time resulted in failure, catastrophe, or death.

German translation: Kandidat B ist ein Mann, der aufbrach, um den Südpol viermal zu erreichen, drei Mal landete er als Leiter und jedes Mal endete es in Misserfolg, Katastrophe oder Tod.

Russian translation: \textcyr{Кандидат B - мужчина , который отправился в путь к Антарктике четыре раза, три раза это был он , кто руководил, и каждый раз это заканчивалось неудачей, катастрофой или смертью.}

Italian translation: Candidato B è un uomo che partì per l'Antartico quattro volte, tre delle quali fu l'uomo al comando, e ogni volta il risultato fu un fallimento, una catastrofe o la morte.

French translation: Candidat B est un homme qui a entrepris une expédition vers l'Antarctique à quatre reprises, trois fois il était à la tête de l'expédition, et chaque fois cela s'est soldé par un échec, une catastrophe ou la mort.

Spanish translation: El candidato B es un hombre que partió hacia la Antártida cuatro veces, tres veces como hombre a cargo, y cada vez resultó en un fracaso, una catástrofe o la muerte.

\vspace{20pt}
English source: But in reality, we often trick ourselves into hiring Candidate B or someone like him.

German translation: Aber, in Wirklichkeit, tun wir uns oft selbst einen Gefallen, indem wir Kandidat B oder jemanden wie ihn einstellen.

Russian translation: \textcyr{Но, на самом деле, мы часто обманываем самих себя, нанимая кандидата Б или кого-то вроде него.}

Italian translation: Ma, in realtà, spesso inganniamo noi stessi nell'assumere candidati come B o qualcuno simile a lui.

French translation: Mais, en réalité, nous trompons souvent nous-mêmes en embauchant le candidat B ou quelqu'un de semblable.

Spanish translation: Pero, en realidad, a menudo nos engañamos al contratar al candidato B o a alguien como él.

\vspace{20pt}
English source: Meanwhile, Candidate A, the Norwegian Roald Amundsen, by any metric, the most successful polar explorer to have ever lived, has been largely forgotten.

German translation: Inzwischen, Kandidat A, der Norweger, ähnlich wie Amundsen, nach jeder Messlatte, der erfolgreichste Polarforscher , der je gelebt hat, wurde größtenteils vergessen.

Russian translation: \textcyr{Между тем, кандидат А, норвежец Роальд Амундсен, по любому критерию, самый успешный полярный исследователь , когда-либо существовавший , был в значительной степени забыт.}

Italian translation: Nel frattempo, il candidato A, il norvegese Roald Amundsen, secondo ogni parametro, il più grande esploratore polare di tutti i tempi, è stato largamente dimenticato.

French translation: Pendant ce temps, le candidat A, le Norvégien Roald Amundsen, selon tous les critères, l'explorateur polaire le plus réussi de tous les temps, est largement tombé dans l'oubli.

Spanish translation: Mientras tanto, el candidato A, el noruego Amundsen, según cualquier métrica, el explorador polar más exitoso que haya vivido jamás, ha sido en gran medida olvidado.

\vspace{20pt}
English source: I did a quick search in my university's library catalog before this talk, and I found no fewer than 26 books that celebrate Shackleton's leadership qualities.

German translation: Ich habe eine schnelle Suche im Bibliothekskatalog meiner Universität durchgeführt, bevor ich hierher kam, und fand nicht weniger als 26 Bücher, die Shackletons Führungsqualitäten feiern.

Russian translation: \textcyr{Я быстро поискал в библиотеке нашего университета перед этим докладом, и я нашел ни меньше , чем 26 книг, которые прославляют лидерство Шеклтона.}

Italian translation: Ho fatto una ricerca rapida nel catalogo della biblioteca universitaria prima di questo intervento e ho trovato non meno di 26 libri che celebrano le qualità di leadership di Shackleton.

French translation: J'ai fait une recherche rapide dans le catalogue de la bibliothèque de mon université avant cette conférence, et j'ai trouvé pas moins de 26 livres qui célébrent les qualités de leadership de Shackleton.

Spanish translation: Hice una búsqueda rápida en el catálogo de la biblioteca de mi universidad antes de esta charla y encontré no menos de 26 libros que celebran las cualidades de liderazgo de Shackleton.


\section{Details on calculating the RTF}
\label{rtf}

The RTF values reported in Table \ref{tab:param_RTF} were obtained by running our model on \textsc{TED-TST-2-2024} with the parameter settings needed to achieve a LAAL of approximately 2000. Specifically:

\vspace{10pt}
\textbf{NAIST}
\vspace{10pt}

\texttt{source-segment-size 600-950}

\texttt{la-n 2}

\texttt{beam 5}

\vspace{10pt}
\textbf{FBK}
\vspace{10pt}

\texttt{extract-attn-from-layer 3}

\texttt{frame-num 2}

\texttt{attn-threshold 0.2-0.4}

\vspace{10pt}
\textbf{SeamlessStreaming}
\vspace{10pt}

\texttt{source-segment-size 400}

\texttt{decision-threshold 0.6-0.9}

\vspace{10pt}
\textbf{TransLLaMa}
\vspace{10pt}

\texttt{wait-k 1}

\texttt{min-read-time 1.2-1.8}

\texttt{asr-model whisper-large-v2}

\vspace{10pt}
\textbf{Ours}
\vspace{10pt}

\texttt{min-read-time 1.2-1.8}

\texttt{asr-model whisper-small.en}

\vspace{10pt}
All the runs were on the same hardware as mentioned in the main text. The RTF was computed as the ratio of (wall) time it took the model to complete translation of the given dataset to the total duration of the corresponding source audio clips.

\section{Dataset statistics}
\label{data_stats}

\begin{table}[h]
\centering
\begin{tabular}{lrrll}
\toprule
Dataset name & $N$ \\
\midrule
\textsc{TED-TST-2023}\tablefootnote{\texttt{https://github.com/RomanKoshkin/transllama}} & 102 \\
\textsc{TED-TST-2024} & 478  \\
\textsc{Fleurs}\tablefootnote{\texttt{https://huggingface.co/datasets/google/fleurs/blob/main/data/en\_us/test.tsv}} & 642  \\
\textsc{AmbiEval} & 96 \\
\bottomrule
\end{tabular}
\caption{Number of samples ($N$).}
\label{tab:dataset_stats}
\end{table}

\end{document}